\documentclass{article}

\usepackage{PRIMEarxiv}

\usepackage[utf8]{inputenc} 
\usepackage[T1]{fontenc}    
\usepackage{hyperref}       

\usepackage{amsmath}
\usepackage{amssymb}
\usepackage{mathtools}
\usepackage{amsthm}
\usepackage{algorithm}
\usepackage{algorithmic}
\usepackage{multirow}
\usepackage{graphicx}
\usepackage{lscape}

\usepackage[capitalize,noabbrev]{cleveref}

\newcommand{\RETURN}{\STATE \textbf{return}\quad}
\newcommand{\R}{\mathbb{R}}

\title{Achieving Constraints in Neural Networks: A Stochastic Augmented Lagrangian Approach
}

\author{
  Diogo Lavado, Cláudia Soares \\
  NOVA School of Science and Technology \\
  Lisbon\\
  \texttt{\{d.lavado,claudia.soares\}@fct.unl.pt} 
   \And
  Alessandra Micheletti \\
  University of Milan,\\
  Milan\\
  \texttt{alessandra.micheletti@unimi.it} \\
}

\begin{document}
\maketitle

\begin{abstract}
  Regularizing Deep Neural Networks (DNNs) is essential for improving generalizability and preventing overfitting. Fixed penalty methods, though common, lack adaptability and suffer from hyperparameter sensitivity. In this paper, we propose a novel approach to DNN regularization by framing the training process as a constrained optimization problem. Where the data fidelity term is the minimization objective and the regularization terms serve as constraints. Then, we employ the Stochastic Augmented Lagrangian (SAL) method to achieve a more flexible and efficient regularization mechanism.
Our approach extends beyond black-box regularization, demonstrating significant improvements in white-box models, where weights are often subject to hard constraints to ensure interpretability.
Experimental results on image-based classification on MNIST, CIFAR10, and CIFAR100 datasets validate the effectiveness of our approach. SAL consistently achieves higher Accuracy while also achieving better constraint satisfaction, thus showcasing its potential for optimizing DNNs under constrained settings.
\end{abstract}

\section{Introduction}

Deep Neural Networks (DNNs) have shown remarkable success in diverse applications. However, overfitting remains a challenge, which calls for effective regularization techniques. 
%
Fixed penalty (FP) methods, like \textit{L}$_1$ or \textit{L}$_2$ regularization, are commonly used but lack adaptability since different layers in DNNs may require different amounts of regularization, FP methods induce the same amount of regularization to every layer.
In addition, the effectiveness of fixed penalties often depends on hyperparameter tuning, specifically of the regularization coefficients, which can not only be challenging and sensitive to the dataset and network architecture but also make it difficult to explore various parameters due to the time-consuming nature of training DNNs.

The Augmented Lagrangian method (ALM) is an optimization technique designed to solve constrained optimization problems. By viewing DNN training as a constrained optimization problem, we can leverage the Augmented Lagrangian method to enforce regularization as constraints.
By iteratively updating the Lagrange multipliers and the penalty parameters, the Augmented Lagrangian method dynamically adapts regularization strengths during the training process. This adaptability allows for a better balance between preventing overfitting and retaining model performance.
Adopting ALM to train neural networks under constraints has been widely used, namely,~\cite{nandwani2019primal} showcased the power of ALM by demonstrating state-of-the-art performance in three NLP benchmarks using a constrained formulation. ~\cite{sangalli2021constrained} harnessed ALM to address class-imbalanced binary classification, while~\cite{fioretto2020predicting,fioretto2021lagrangian} applied it to optimal power flow prediction problems and energy domains. 
For problems involving partial differential equations (PDEs),~\cite{hwang2021lagrangian} introduced an ALM approach to enforce physical conservation laws of kinetic PDEs on neural networks. Additionally,~\cite{son2023enhanced} proposed an ALM strategy to train Physically informed neural networks (PINNs) by deriving a novel sequence of loss functions with adaptively balanced loss components.

The Alternating Direction Method of Multipliers (ADMM), introduced and popularized by the work of~\cite{boyd2011distributed}, has emerged as a powerful tool for statistics and machine learning challenges involving numerous features or training examples. ADMM adopts a decomposition-alternating approach, where solutions to smaller local subproblems are harmonized to derive a solution to the larger global problem. 
By combining the benefits of dual decomposition and augmented Lagrangian methods, ADMM excels in handling combinatorial constraints and supports efficient parallelization.
Various applications use ADMM to train DNNs under constraints, for instance,~\cite{zhao2018admm,zhao2019admm} developed an ADMM-based framework against adversarial attacks, while ~\cite{ye2018progressive} used ADMM for weight pruning in DNNs. 

In the context of white-box models, constrained optimization introduces a delicate balance between interpretability and performance. Unlike black-box DNNs, white-box models often involve hard constraints, where parameters hold meaningful interpretations within specific feasible sets. Fixed penalty methods prove inadequate for upholding such hard constraints due to their lack of adaptability and fixed balance between data fidelity and penalties.
By formulating DNN training as a constrained optimization problem, SAL and ADMM effectively handle complex hard constraints. The introduction of Lagrange multipliers in SAL and the decomposition-alternating approach in ADMM ensures the enforcement of hard constraints while maintaining model performance.

The contributions of this work are as follows: (1) a novel approach to DNN regularization by formulating the training process as a constrained optimization problem. By leveraging the SAL method, we achieve more flexible and efficient regularization. (2) Our method improves white-box models' performance, ensuring interpretability while effectively handling hard constraints. Experimental results on various datasets validate its effectiveness. (3) We demonstrate the applicability of ADMM for DNN training under constraints.
This work presents a comprehensive and versatile regularization framework, paving the way for insights into constrained optimization and interpretable DNNs.

\section{Theoretical Background}

\subsection{Augmented Lagrangian Method}

Consider a generic optimization problem for an objective function $F: \Theta \to \R$ subject to constraints $C(\theta) = [c_1(\theta), \ldots, c_m(\theta) ]$:
\begin{align}~\label{eq:OrignOptProblem}
    \arg\min_{\theta \in \Theta} F(\theta); 
    \quad \text{s.t.}  \; C(\theta) = 0.
\end{align}

The augmented Lagrangian method (ALM)~\cite{gabay1976dual} relaxes the problem in~\ref{eq:OrignOptProblem} into an unconstrained optimization problem. 
Specifically, it harmonizes two earlier methods, the quadratic penalty method and 
the method of Lagrangian multipliers, which suffer from training instability and non-convergence due to the difficulty of convexifying loss functions.
ALM augments the Lagrangian function with a quadratic term that helps drive the solution toward the constraint, thus forming the augmented Lagrangian function:

\begin{align}\label{eq:ALMFunction}
L_{\rho}(\theta, \lambda) &= F(\theta) + \big<\lambda, C(\theta)\big> + \frac{\rho}{2} \Vert C(\theta) \Vert_2^2,
%
\end{align}
where $\lambda$ is the Lagrangian multiplier vector and $\rho$ is the penalty coefficient that controls the trade-off between the objective function and the constraint violation.
Similarly to the Fixed Penalty (FP) method, if $\rho$ is chosen to be too large, then the optimization problem can become stiff and convergence will be very slow.
If chosen to be too small, the solution will deviate from the feasible space enforced by the constraints.

The augmented Lagrangian method solves problem (\ref{eq:OrignOptProblem}) by executing the following recursion in $k$

\begin{align}\label{eq:auglag_opt_prob}
\min_{\theta \in \Omega} L(\theta^k, \lambda^k),
\end{align}

which can be translated into the following steps:

\begin{align}
    &\theta^{k +1} \gets \arg\min_{\theta \in \Omega} L_{\rho}(\theta, \lambda^{k}),\label{eq:auglag_optstep}\\
     &\lambda^{k + 1} \gets \lambda^{k} + \rho C(\theta^{k +1})\label{eq:auglag_multstep}
\end{align}

The optimization step on (\ref{eq:auglag_optstep}) computes the minimum of the augmented Lagrangian cost function with respect to $\theta$, and the update step on (\ref{eq:auglag_multstep}) updates the Lagrangian multipliers based on the error in the constraints.
The convergence of ALM does not depend on the choice of $\rho$. A large $\rho$ leads to faster convergence in terms of needed iterations, however, each iteration becomes more difficult to compute because the optimization step (\ref{eq:auglag_optstep}) becomes more ill-conditioned. Thus, it becomes crucial to find a penalty that balances these objectives of fast convergence and well-conditioned minimization.

\begin{algorithm}[ht]
    \caption{Deterministic Augmented Lagrangian Method}
    \label{alg:aug_lag}
    \begin{algorithmic}[1]
        \STATE Initialize $\theta^0, \lambda^0 \gets 0$
        \FOR{$k = 0, 1, \dots, K - 1$}
            \STATE $\theta^{k+1} \gets \arg\min_{\theta} L_{\rho}(\theta, \lambda^{k})$ \label{alg:aug_lag_step1}
            \STATE $\lambda^{k+1} \gets \lambda^{k} + \rho C(\theta^{k+1})$ \label{alg:aug_lag_step2}
        \ENDFOR
    \end{algorithmic}
\end{algorithm}

\subsection{Alternating Direction Method of Multipliers}
The Alternating Direction Method of Multipliers (ADMM) \cite{gabay1976dual,boyd2011distributed} is particularly effective for solving problems that can be decomposed into smaller, simpler subproblems, such as those arising in machine learning and signal processing.
ADMM splits the primal variables into two sets, one with respect to the primal function $F$ and the other to the constraints. Thus, we cast the problem in (\ref{eq:OrignOptProblem}) into a new problem via variable splitting:

\begin{align}\label{eq:admm_valsplit_opt_prob}
\arg\min_{\theta \in \Omega, \mu \in \Omega}& \quad F(\theta) 
\ + C(\mu)\\
\text{s.t.} & \quad \theta\nonumber = \mu.
\end{align}

The augmented Lagrangian function for (\ref{eq:admm_valsplit_opt_prob}) is now defined as:

\begin{align}\label{eq:admm_Lagrangian}
L_{\rho}(\theta, \mu, \lambda) &= F(\theta) + C(\mu) + \big<\lambda, (\theta - \mu)\big> + \frac{\rho}{2} \Vert \theta - \mu \Vert_2^2\nonumber
\\
&= F(\theta) + C(\mu) + \frac{\rho}{2} \Vert \theta - \mu + u\Vert_2^2,
\end{align}

where $u = \lambda / \rho$ is the scaled Lagrangian multiplier vector.
To determine $u$, we follow the strategy of the augmented Lagrangian algorithm with the following iteration

\begin{align}
    (\theta^{k +1}, \mu^{k +1}) &\gets \arg\min_{(\theta,\mu)} L_{\rho}(\theta, \mu, \lambda^{k})\label{eq:admm_optstep}\\
     \lambda^{k + 1} &\gets \lambda^{k} + \rho (\theta^{k +1} - \mu^{k +1})\label{eq:admm_multstep}
\end{align}

Then, the method of alternating minimization lets us achieve the following equivalent algorithm

\begin{align}
    \theta^{k +1} &\gets \arg\min_{\theta}  F(\theta) + \frac{\rho}{2} \Vert \theta - \mu^{k} + u^k\Vert_2^2\label{eq:admm_x_optstep}\\
    \mu^{k +1} &\gets \arg\min_{\mu}  C(\mu) + \frac{\rho}{2} \Vert \theta^{k + 1} - \mu + u^k\Vert_2^2\label{eq:admm_y_optstep}\\
     u^{k + 1} &\gets u^{k} + \rho (\theta^{k +1} - \mu^{k +1})\label{eq:admm_multstep_2}.
\end{align}

Since alternating minimization can be seen as a form of coordinated descent~\cite{wright2015coordinate}, this algorithm converges to the desired solution of equation (\ref{eq:admm_valsplit_opt_prob}).

\section{Related Work}

\subsection{Fixed Penalty Method}

Fixed penalty (FP) methods attempt to convert the original optimization problem in (\ref{eq:OrignOptProblem}) into an unconstrained optimization problem by augmenting the loss function with a penalty term to handle constraints:

\begin{align}\label{eq:fixed_penalty_opt_prob}
\min_{\theta \in \Omega} & \quad F(\theta) + \rho \Vert C(\theta) \Vert_2^2,
\end{align}

where $F$ denotes the loss function, $\theta$ represents the trainable parameters of the model, and $\rho > 0$ serves as the penalty parameter, balancing the trade-off between data fidelity and constraint enforcement during the training process.
However, the effectiveness of this method relies heavily on manual fine-tuning of the penalty parameter $\rho$, as an inappropriate value can lead to an unstable optimization process. Moreover, the FP approach lacks adaptability since it induces the same amount of regularization across all layers of the DNN, regardless of their specific requirements. This lack of adaptability can hinder its performance on complex tasks with varying regularization needs. Additionally, the FP method does not guarantee constraint satisfaction below a certain threshold of interest, $\Vert C(\theta) \Vert_2^2 \leq \epsilon$. This limitation further motivates the search for more robust and efficient regularization methods for training DNNs.

\subsection{Stochastic Augmented Lagrangian (SAL)}

The Stochastic Augmented Lagrangian (SAL) applies the Augmented Lagrangian Method (ALM) to train neural networks and has garnered attention for its effectiveness in handling constrained optimization problems.
\cite{nandwani2019primal} demonstrated the power of SAL in achieving state-of-the-art performance in three Natural Language Processing (NLP) benchmarks, while \cite{sangalli2021constrained} utilized SAL to address class-imbalanced binary classification, improving performance on imbalanced datasets. In energy domains,~\cite{fioretto2020predicting,fioretto2021lagrangian} successfully applied SAL to optimal power flow prediction problems. In the context of Partial Differential Equations (PDEs),~\cite{hwang2021lagrangian} employed SAL to enforce physical conservation laws of kinetic PDEs on neural networks. Additionally,~\cite{son2023enhanced} proposed a SAL strategy for training Physically Informed Neural Networks (PINNs) with adaptively balanced loss components.

Although these approaches have achieved state-of-the-art results in their respective domains, we recognize that the use of stochastic mini-batch gradients in machine learning problems calls for modifications to the original ALM algorithm. In our work, we draw inspiration from the research of~\cite{dener2020training}, where they address this setting by designing a SAL algorithm that adapts to stochastic mini-batch gradients. While~\cite{dener2020training} applied their SAL strategy to train DNNs under physical constraints, our work aims to extend and optimize this approach for training DNNs under general constraints and regularization.

\subsection{Stochastic ADMM}

Stochastic ADMM (S-ADMM) extends the traditional ADMM algorithm to handle large-scale optimization problems with noisy data and a large number of constraints and variables. In S-ADMM, the primal and dual variables are updated following the traditional ADMM strategy, i.e., the steps (\ref{eq:admm_optstep}) and (\ref{eq:admm_multstep}), but with an approximated augmented Lagrangian function given by~\cite{ouyang2013stochastic,suzuki2013dual}:

\begin{align}
\hat{L}_{\rho,k}(\theta, \mu, u) = &F(\theta^k; \xi^{k+1}) + \partial F(\theta^k; \xi^{k+1})^T(\theta - \theta^k) + \\
&C(\mu) + \frac{\rho}{2} \Vert \theta - \mu + u\Vert_2^2 + \frac{\Vert \theta - \theta^k \Vert^2}{2\eta^{k}}.\nonumber
\end{align}

Here, $\xi^{k+1}$ is a random sample drawn from an unknown distribution, $\rho$ is the penalty parameter, and $\eta^{k+1}$ is the step size. The S-ADMM algorithm is particularly useful when dealing with complex and large-scale optimization problems involving noisy data.

\section{Methodology}

\subsection{Stochastic Augmented Lagrangian (SAL)}

Augmented Lagrangian methods typically adopt a two-level nested loop structure. The inner problem (\ref{eq:auglag_optstep}) is solved using conventional unconstrained optimization methods, while the outer loop updates the Lagrangian multipliers and the penalty factor based on the constraint violation of the inner loop solution.
In their work, the authors of~\cite{dener2020training} propose a Stochastic Augmented Lagrangian (Aug-Lag) method tailored for DNN training presented in Algorithm~\ref{alg:sal_method}. The inner loop of Aug-Lag involves using SGD to solve the unconstrained optimization problem. Unlike traditional ALMs that rely on dynamic convergence tolerances, Aug-Lag iterates through the entire training dataset once per outer iteration.
In the outer loop, Aug-Lag accepts updates to the Lagrange multipliers whenever the SGD solution achieves a sufficient decrease in the constraint violation. The multiplier update remains unchanged from the conventional aug-Lag method. The penalty parameter increases with a fixed factor when the SGD solution fails to satisfy the sufficient decrease criteria.
Lastly, Aug-Lag makes use of learning rate decay, but it is reset to its initial value in each outer iteration. This ensures independent decay rates for each constructed training subproblem, preventing the model from stagnating.

\begin{algorithm}[ht]
    \caption{Stochastic Augmented Lagrangian Method for Constrained ML Training~\cite{dener2020training}}
    \label{alg:sal_method}
    \begin{algorithmic}[1]
        \REQUIRE Update tolerance $\eta$, convergence tolerances $\epsilon_f$ and $\epsilon_c$, initial penalty $\mu_{\text{init}}$, penalty update factor $\sigma$, penalty safeguard $\mu_{\text{max}}$, batch size $n_{\text{batch}}$, number of random shuffles $n_{\text{shuffle}}$, and number of Augmented Lagrangian iterations per shuffle $n_{\text{aug-Lag}}$.
        \RETURN Trained model parameters $\theta*$
        \STATE $C_{\text{best}} \gets C(\theta_0)$
        \STATE $\lambda_0 \gets 0$
        \FOR{$\text{shuffle} = 0, 1, 2, \ldots, n_{\text{shuffle}}$}
            \STATE Randomly shuffle training dataset, split into batches of size $n_{\text{batch}}$
            \STATE $\mu_0 \gets (\text{shuffle} + 1) \times \mu_{\text{init}}$
            \FOR{$k = 0, 1, 2, \ldots, n_{\text{aug-Lag}} - 1$}
                \STATE $\theta_{k+1} \gets \arg\min_\theta L_\rho(\theta_k, \lambda_k)$
                \IF{$\|C(\theta_{k+1})\|_2^2 \leq \eta \|C_{\text{best}}\|_2^2$}
                    \IF{$F(\theta_{k+1}) \leq \epsilon_f$ and $\|C(\theta_{k+1})\|_2^2 \leq \epsilon_c$}
                        \STATE Terminate with $\theta^* = \theta_{k+1}$
                    \ENDIF
                    \STATE $\lambda_{k+1} \gets \lambda_k + \mu_k C(\theta_{k+1})$
                    \STATE $C_{\text{best}} \gets C(\theta_{k+1})$
                    \STATE $\mu_{k+1} \gets \mu_k$
                \ELSE
                    \STATE $\lambda_{k+1} \gets \lambda_k$
                    \STATE $\mu_{k+1} \gets \min(\sigma \times \mu_k, \mu_{\text{max}})$
                \ENDIF
            \ENDFOR
            \STATE $\lambda_0 \gets \lambda_{n_{\text{aug-Lag}}}$
        \ENDFOR
    \end{algorithmic}
\end{algorithm}

\subsection{Constraints in Neural Networks}

Deep Neural networks are often subject to constraints in order to impose certain conditions on the parameters of the model or to ensure desirable properties. Constraints in neural networks can be broadly categorized into hard and soft constraints.
%
Hard constraints are strict conditions that must be satisfied for the model to be valid. Violating hard constraints typically renders the model unsuitable for the task, this is of particular importance in white-box models where parameters are meaningful under specific domain values. On the other hand, soft constraints are more flexible and allow some level of violation. For instance, $L_1$ regularization is a soft constraint since we do not want a model to strictly satisfy it, rather it helps the model traverse the loss landscape in a better direction

Several types of constraints can be imposed on neural networks, we focused on the following:

\textbf{$L_1$ and $L_2$ Norm Constraints:} $L_1$ and $L_2$ norm constraints are commonly used to control the magnitude of model parameters. $L_1$ norm constraints, $\Vert \boldsymbol{w} \Vert_1$, enforce sparsity by encouraging many parameters to be exactly zero. $L_2$ norm constraints, $\Vert \boldsymbol{w} \Vert_2^2$, promote weight decay, effectively penalizing large parameter values. Both norms help prevent overfitting and can lead to more robust models.

\textbf{Orthogonality Constraints:} Orthogonality constraints enforce that certain weight matrices are orthogonal~\cite{wang2020orthogonal}:

\begin{align}
\Vert \boldsymbol{W}^\top \boldsymbol{W} - \boldsymbol{I} \Vert_F^2
\end{align}

where $\boldsymbol{W}$ is the weight matrix and $\boldsymbol{I}$ is the identity matrix. This property is often beneficial in tasks involving feature extraction or when interpretability is essential.

\textbf{Non-Negativity Constraints:} Non-negativity constraints ensure that parameters remain non-negative throughout the training process. This is particularly useful in scenarios where negative values are not meaningful, such as distribution parameters.

\begin{align}
(-w_i)_+, \text{ for all } i,
\end{align}

where $(h)_+ = \max(0, h)$ is the Rectified Lineat Unit (ReLU) and $w_i$ represents a model parameter.

\subsection{Constraint Violation Metric}

We introduce a metric designed to gauge constraint violation within DNNs. The purpose of this metric is to quantitatively evaluate the degree to which the specified constraints are breached by the model. We focus on assessing the constraint violation at the inception of the DNN and then tracing its evolution as training progresses.
Given a set of constraints $\mathcal{C}$ imposed on the DNN, we present the constraint violation metric as follows:
Let $\boldsymbol{\theta}$ represent the trainable parameters of the DNN at the commencement of training, and $C(\boldsymbol{\theta})$ indicate the vector of constraint violation values for each constraint within $\mathcal{C}$. The constraint violation metric is defined as:

\begin{equation}
\text{Constraint Violation (CV)} = \left( \sum_{i=1}^{m} |C_i(\boldsymbol{\theta})|^p \right)^{\frac{1}{p}},
\end{equation}

where $m$ is the number of constraints within $\mathcal{C}$, and $p$ stands as a hyperparameter controlling the metric's sensitivity.
The works of~\cite{dener2020training} and~\cite{fioretto2021lagrangian} set $p=2$ to assess constraint violation. While this is suitable for hard constraints, alternative $L_p$ norms might be better suited for certain soft constraints. For instance, $L_1$ regularization shouldn't be strictly satisfied by a DNN; therefore, $L_\infty$ could be a more appropriate choice for such a constraint. This is especially pertinent when different architectures exhibit varying parameter counts, rendering the comparison of constraint violation measures unfeasible.

In order to measure the improvement of the constraint violation across different architectures and stages, we introduce an extension of the metric:

\begin{align}
    \text{CV(e)} = CV_e / CV_0, 
\end{align}
where $CV_i$ signifies the constraint violation at step/epoch $i$, and as such, CV($i$) denotes the progression of constraint violation relative to the initial CV prior to training.
Thus, the constraint violation metric proposed here is designed to provide a summary of the adherence of DNNs to the imposed constraints. The $L_p$ norm allows us to control the sensitivity of the metric, making it adaptable to different types of constraints and model architectures.

\section{Experiments}

\subsection{Experimental Setup}

Our experimentation includes diverse neural network architectures, such as ResNet13~\cite{he2016deep}, VGG11~\cite{simonyan2014very}, CNN, and GENEOnet~\cite{bergomi2019towards}. These architectures underwent evaluations under various constraints, including $L_1$ and $L_2$ regularization, orthogonality, and non-negativity. We examined datasets like CIFAR-10, CIFAR-100, and MNIST for image classification.
The datasets were augmented through isomorphic transformations (i.e., translations, rotations, and inversions), with data normalized as a preprocessing step.
The data fidelity function $F$ used was the cross-entropy loss.
Utilizing a batch size of 128, we explored learning rates ranging from 0.001 to 0.00001, employing Adam and SGD optimizers. Additionally, for white-box models, the L-BFGS optimizer was employed, capitalizing on their limited parameter count.
Our methodology encompassed different training iterations, where we contrasted the fixed penalty method (FP) with different penalty coefficient combinations $\rho_{i} \in[0.0001, 0.1]$, against the stochastic augmented Lagrangian method (SAL) outlined in Algorithm~\ref{alg:sal_method} and stochastic ADMM (S-ADMM)~\cite{ouyang2013stochastic}. The penalty parameter $\rho$ in SAL and S-ADMM varied in each experiment, although this variation did not influence method performance, given the progressive adaptation of the penalty during training, corroborated by~\cite{dener2020training}.
Finally, our code is openly accessible and seamlessly integratable with existing architectures. Both the SAL and S-ADMM implementations are programmed using PyTorch and work as wrappers for data fidelity losses. The formulation of constraints is straightforward, rendering the incorporation of SAL and S-ADMM into ongoing projects a straightforward process.

\subsection{Numerical Results}

\begin{table}[t]
\caption{Quantitative results of CNN and GENEOnet on the MNIST Dataset. CV(-1) indicates the constraint violation at the end of training relative to the initial network initialization, while CV$_{p=2}$ represents the constraint violation $L_2$ norm at test time. Evaluation of other DNNs was omitted due to the problem's simplicity for networks like VGG11 and ResNet13. The reported results are the averages across 100 runs for each method, encompassing different hyperparameter configurations.
}
\centering
\resizebox{\columnwidth}{!}{%
\begin{tabular}{llllll}
\hline
\multicolumn{1}{c}{DNN} &
  \multicolumn{1}{c}{Constraints} &
  \multicolumn{1}{c}{Method} &
  \multicolumn{1}{c}{CV(-1)} &
  \multicolumn{1}{c}{CV$_{p=2}$} &
  \multicolumn{1}{c}{Accuracy} \\ \hline
\multicolumn{1}{l|}{\multirow{3}{*}{CNN}} &
  \multicolumn{1}{l|}{\multirow{3}{*}{\begin{tabular}[c]{@{}l@{}}$L_2$;\\ Orthogonality;\end{tabular}}} &
  \multicolumn{1}{l|}{FP} &
  \textbf{0.72 ($\sigma$ 0.06)} &
  3.2K ($\sigma$ 7.6) &
  0.70 ($\sigma$ 0.02) \\
\multicolumn{1}{l|}{} &
  \multicolumn{1}{l|}{} &
  \multicolumn{1}{l|}{SAL} &
  0.73 ($\sigma$ 0.03) &
  \textbf{2.0K ($\sigma$ 15.4)} &
  \textbf{0.75 ($\sigma$ 0.01)} \\
\multicolumn{1}{l|}{} &
  \multicolumn{1}{l|}{} &
  \multicolumn{1}{l|}{S-ADMM} &
  0.73 ($\sigma$ 0.04) &
  2.7K ($\sigma$ 115) &
  0.74 ($\sigma$ 0.01) \\ \hline
\multicolumn{1}{l|}{\multirow{3}{*}{GENEOnet}} &
  \multicolumn{1}{l|}{\multirow{3}{*}{\begin{tabular}[c]{@{}l@{}}$L_2$;\\ Non-Negativity; \\ Orthogonality;\end{tabular}}} &
  \multicolumn{1}{l|}{FP} &
  0.72 ($\sigma$ 0.07) &
  2.7K ($\sigma$ 152) &
  0.80 ($\sigma$ 0.1) \\
\multicolumn{1}{l|}{} &
  \multicolumn{1}{l|}{} &
  \multicolumn{1}{l|}{SAL} &
  0.74 ($\sigma$ 0.04) &
  2.4K ($\sigma$ 30.1) &
  0.82 ($\sigma$ 0.01) \\
\multicolumn{1}{l|}{} &
  \multicolumn{1}{l|}{} &
  \multicolumn{1}{l|}{S-ADMM} &
  \textbf{0.72 ($\sigma$ 0.03)} &
  \textbf{1.7K ($\sigma$ 103)} &
  \textbf{0.83 ($\sigma$ 0.02)} \\ \hline
\end{tabular}%
}
\label{tab:mnist}
\end{table}

\begin{table}[t]
\caption{Quantitative results of ResNet13 and VGG11 on CIFAR100, where CV(-1) indicates the constraint violation at the end of training relative to the initial network initialization, while CV$_{p=2}$ represents the constraint violation $L_2$ norm at test time. Evaluation of other DNNs was omitted due to the problem's simplicity for networks like VGG11 and ResNet13. The reported results are the averages across 60 runs for each method, encompassing different hyperparameter configurations.}
\centering
\resizebox{\columnwidth}{!}{%
\begin{tabular}{llllll}
\hline
\multicolumn{1}{c}{DNN} &
  \multicolumn{1}{c}{Constraints} &
  \multicolumn{1}{c}{Method} &
  \multicolumn{1}{c}{CV(-1)} &
  \multicolumn{1}{c}{CV$_{p=2}$} &
  \multicolumn{1}{c}{Accuracy} \\ \hline
\multicolumn{1}{l|}{\multirow{3}{*}{ResNet13}} &
  \multicolumn{1}{l|}{\multirow{3}{*}{\begin{tabular}[c]{@{}l@{}}$L_1$;\\ Orthogonality;\end{tabular}}} &
  \multicolumn{1}{l|}{FP} &
  \textbf{0.11 ($\sigma$ 0.04)} &
 \textbf{ 330.5 ($\sigma$ 1.4)} &
  0.72 ($\sigma$ 0.02) \\
\multicolumn{1}{l|}{} & \multicolumn{1}{l|}{} & \multicolumn{1}{l|}{SAL}    & 
0.14 ($\sigma$ 0.02)         & 
363.4 ($\sigma$ 30.1)  & 
\textbf{0.77 ($\sigma$ 0.01)} \\
\multicolumn{1}{l|}{} & \multicolumn{1}{l|}{} & \multicolumn{1}{l|}{S-ADMM} &
0.13 ($\sigma$ 0.01)         & 
346 ($\sigma$ 31.9)          & 
0.75 ($\sigma$ 0.01)          \\ \hline
\multicolumn{1}{l|}{\multirow{3}{*}{VGG11}} &
  \multicolumn{1}{l|}{\multirow{3}{*}{\begin{tabular}[c]{@{}l@{}}$L_1$;\\ Orthogonality;\end{tabular}}} &
  \multicolumn{1}{l|}{FP} &
  0.467 ($\sigma$ 0.002) &
  295.3 ($\sigma$ 1.4) &
  0.72 ($\sigma$ 0.03) \\
\multicolumn{1}{l|}{} & \multicolumn{1}{l|}{} & \multicolumn{1}{l|}{SAL}    & 
\textbf{0.461 ($\sigma$ 0.001)}         & 
\textbf{294.2 ($\sigma$ 0.85)}  & 
\textbf{0.73 ($\sigma$ 0.02)} \\
\multicolumn{1}{l|}{} & \multicolumn{1}{l|}{} & \multicolumn{1}{l|}{S-ADMM} & 
0.47 ($\sigma$ 0.01)         & 
296 ($\sigma$ 0.92)          & 
0.72 ($\sigma$ 0.01)          \\ \hline
\end{tabular}%
}
\label{tab:cifar100}
\end{table}

\begin{table}[]
\caption{Quantitative results of CNN, GENEOnet, ResNet13, and VGG11 on the CIFAR10 dataset. 
CV(-1) indicates the constraint violation at the end of training relative to the initial network initialization, while CV$_{p=2}$ represents the constraint violation $L_2$ norm at test time. Evaluation of other DNNs was omitted due to the problem's simplicity for networks like VGG11 and ResNet13. The reported results are the averages across 50 runs for each method, encompassing different hyperparameter configurations.}
\centering
\resizebox{\columnwidth}{!}{%
\begin{tabular}{llllll}
\hline
\multicolumn{1}{c}{DNN} &
  \multicolumn{1}{c}{Constraints} &
  \multicolumn{1}{c}{Method} &
  \multicolumn{1}{c}{CV(-1)} &
  \multicolumn{1}{c}{CV$_{p=2}$} &
  \multicolumn{1}{c}{Accuracy} \\ \hline
\multicolumn{1}{l|}{\multirow{3}{*}{CNN}} &
  \multicolumn{1}{l|}{\multirow{3}{*}{\begin{tabular}[c]{@{}l@{}}$L_2$;\\ Orthogonality;\end{tabular}}} &
  \multicolumn{1}{l|}{FP} &
  0.19 ($\sigma$ 0.06) &
  27.5 ($\sigma$ 9.3) &
  0.36 ($\sigma$ 0.01) \\
\multicolumn{1}{l|}{} &
  \multicolumn{1}{l|}{} &
  \multicolumn{1}{l|}{SAL} &
  \textbf{0.12 ($\sigma$ 0.01)} &
  \textbf{16.9 ($\sigma$ 1.7)} &
  \textbf{0.44 ($\sigma$ 0.007)} \\
\multicolumn{1}{l|}{} &
  \multicolumn{1}{l|}{} &
  \multicolumn{1}{l|}{S-ADMM} &
  0.21 ($\sigma$ 0.02) &
  45.1 ($\sigma$ 3.2) &
  0.41 ($\sigma$ 0.02) \\ \hline
\multicolumn{1}{l|}{\multirow{3}{*}{GENEOnet}} &
  \multicolumn{1}{l|}{\multirow{3}{*}{\begin{tabular}[c]{@{}l@{}}$L_2$;\\ Orthogonality;\\ Non-Negativity;\end{tabular}}} &
  \multicolumn{1}{l|}{FP} &
  \textbf{0.72 ($\sigma$ 0.02)} &
  \textbf{4.2K ($\sigma$ 996)} &
  0.123 ($\sigma$ 0.01) \\
\multicolumn{1}{l|}{} &
  \multicolumn{1}{l|}{} &
  \multicolumn{1}{l|}{SAL} &
  0.73 ($\sigma$ 0.01) &
  4.4K ($\sigma$ 412) &
  \textbf{0.128 ($\sigma$ 0.02)} \\
\multicolumn{1}{l|}{} &
  \multicolumn{1}{l|}{} &
  \multicolumn{1}{l|}{S-ADMM} &
  0.82 ($\sigma$ 0.06) &
  4.6K ($\sigma$ 1K) &
  0.125 ($\sigma$ 0.03) \\ \hline
\multicolumn{1}{l|}{\multirow{3}{*}{ResNet13}} &
  \multicolumn{1}{l|}{\multirow{3}{*}{\begin{tabular}[c]{@{}l@{}}$L_1$;\\ Orthogonality;\end{tabular}}} &
  \multicolumn{1}{l|}{FP} &
  \textbf{0.10 ($\sigma$ 0.02)} &
  \textbf{276 ($\sigma$ 61)} &
  0.71 ($\sigma$ 0.11) \\
\multicolumn{1}{l|}{} &
  \multicolumn{1}{l|}{} &
  \multicolumn{1}{l|}{SAL} &
  0.12 ($\sigma$ 0.002) &
  323 ($\sigma$ 6.6) &
  \textbf{0.88 ($\sigma$ 0.002)} \\
\multicolumn{1}{l|}{} &
  \multicolumn{1}{l|}{} &
  \multicolumn{1}{l|}{S-ADMM} &
  0.12 ($\sigma$ 0.001) &
  308 ($\sigma$ 8.1) &
  0.83 ($\sigma$ 0.006) \\ \hline
\multicolumn{1}{l|}{\multirow{3}{*}{VGG11}} &
  \multicolumn{1}{l|}{\multirow{3}{*}{\begin{tabular}[c]{@{}l@{}}$L_1$;\\ Orthogonality;\end{tabular}}} &
  \multicolumn{1}{l|}{FP} &
  0.46 ($\sigma$ 0.02) &
  339 ($\sigma$ 48) &
  0.62 ($\sigma$ 0.22) \\
\multicolumn{1}{l|}{} &
  \multicolumn{1}{l|}{} &
  \multicolumn{1}{l|}{SAL} &
  0.32 ($\sigma$ 0.01) &
  \textbf{100 ($\sigma$ 12)} &
  \textbf{0.87 ($\sigma$ 0.002)} \\
\multicolumn{1}{l|}{} &
  \multicolumn{1}{l|}{} &
  \multicolumn{1}{l|}{S-ADMM} &
  \textbf{0.31 ($\sigma$ 0.01) }&
  231 ($\sigma$ 18) &
  0.86 ($\sigma$ 0.004) \\ \hline
\end{tabular}%
}
\label{tab:cifar10}
\end{table}

In this section, we conduct a comprehensive evaluation of different Deep Neural Networks (DNNs) across the MNIST, CIFAR10, and CIFAR100 datasets, employing different constraint enforcement methods.
Regarding the MNIST dataset (Table \ref{tab:mnist}), the Stochastic Augmented Lagrangian (SAL) and Stochastic Alternating Direction Method of Multipliers (S-ADMM) consistently outperform the Fixed Penalty (FP) method. SAL, specifically, demonstrates an optimal trade-off between data fidelity and constraint adherence in both the CNN and GENEOnet architectures.
The selected models possess a modest number of parameters (approximately 100K each), and the exclusion of more intricate architectures from testing is due to their suitability for modeling the MNIST challenge. That is, the chosen architectures emphasize the necessity for regularization methods to strike a balance between constraint application and data fidelity.
For the white box model, GENEOnet, the enforcement of non-negativity on its parameters is essential for interpretability. All methods successfully enforce this constraint throughout training. Nonetheless, FP requires a substantial penalty coefficient $\rho_i$ to achieve this, potentially diverting attention from the data fidelity term and leading to suboptimal performance compared to other methods.
On the CIFAR100 dataset (Table \ref{tab:cifar100}) our evaluation encompasses larger networks with millions of parameters. This is crucial because soft constraints such as $L_1$ are applied to a greater number of parameters, introducing a more delicate equilibrium between constraint enforcement and data fidelity within the loss function. 
Nonetheless, SAL consistently achieves the highest Accuracy for both ResNet13 and VGG11, surpassing FP and S-ADMM. 
In this context, SAL does not lead to significant improvements in constraint enforcement over the other methods,
instead, it achieves a balanced trade-off between data-fidelity and constraint enforcement.
Notably, the performance of S-ADMM consistently keeps pace with SAL, demonstrating competitive Accuracy and constraint enforcement. This underscores that, despite a slightly relaxed constraint adherence, S-ADMM effectively achieves competitive predictive performance. In scenarios involving parallel optimization paradigms, S-ADMM emerges as a strong contender and may indeed be the method of choice.

In the case of the CIFAR10 dataset (Table \ref{tab:cifar10}), SAL remains the most effective regularization method in terms of Accuracy, consistently outperforming FP and S-ADMM across all evaluated architectures.
Notably, as we delve into larger networks, both SAL and S-ADMM demonstrate significant performance advantages over the FP method. It's worth highlighting that FP exhibits a higher standard deviation in Accuracy compared to the other methods, suggesting greater instability across different penalty configurations, underscoring the limitations of fixed penalty methods.
%
 SAL excels by securing the top CV$_{p=2}$ score in the VGG11 and CNN models, substantially outperforming FP and S-ADMM. This compelling result, combined with SAL's superior Accuracy, showcases its unique ability to strike an optimal balance between data fidelity and constraint enforcement. Moreover, while FP records the best average CV$_{p=2}$ score in the ResNet13 and GENEOnet architectures, this achievement comes at a noticeable cost in terms of Accuracy. This observation reinforces the notion that SAL excels in finding the optimal equilibrium between constraint imposition and data fidelity, making it a promising choice for various deep learning tasks and architectures.

\section{Conclusions}

This work addresses the imperative need for effective Deep Neural Network (DNN) regularization to counter overfitting and enhance generalization. Conventional fixed penalty methods exhibit limitations in adaptability and sensitivity to hyperparameters. To mitigate these issues, we present an innovative approach that formulates DNN training as a constrained optimization problem, prioritizing data fidelity minimization while treating regularization terms as essential constraints. This conceptual foundation paves the way for the application of the Stochastic Augmented Lagrangian (SAL) method, introducing a dynamic and efficient regularization strategy.
Notably, our approach's benefits extend beyond black-box regularization, demonstrating substantial enhancements in white-box models subject to stringent weight constraints for interpretability.
Empirical validation across diverse datasets, encompassing image classification benchmarks such as MNIST, CIFAR10, and CIFAR100, underscores the efficacy of our methodology. SAL consistently outperforms in terms of Accuracy and constraint enforcement, indicating its potential to optimize DNNs under constraint-driven scenarios.
In essence, our study introduces a pragmatic alternative to fixed penalty methods, emphasizing SAL's adaptability and performance improvement. Empirical results support our approach, indicating improved performance and interpretability in DNNs.

\bibliography{biblio}
\bibliographystyle{unsrt}

\end{document}